\documentclass[Afour,sageh,times]{sagej}

\usepackage{moreverb,url}

\usepackage[colorlinks,bookmarksopen,bookmarksnumbered,citecolor=red,urlcolor=red]{hyperref}
\usepackage{subcaption}
\usepackage{gensymb}

\newcommand\BibTeX{{\rmfamily B\kern-.05em \textsc{i\kern-.025em b}\kern-.08em
T\kern-.1667em\lower.7ex\hbox{E}\kern-.125emX}}

\def\volumeyear{2016}

\begin{document}

\runninghead{Chung et al.}

\title{Pohang Canal Dataset: A Multimodal Maritime Dataset for Autonomous Navigation in Restricted Waters}

\author{Dongha Chung, Jonghwi Kim, Changyu Lee, and Jinwhan Kim}

\affiliation{Department of Mechanical Engineering, KAIST}

\corrauth{Jinwhan Kim, Department of Mechanical Engineering, KAIST}

\email{jinwhan@kaist.ac.kr}

\begin{abstract}
This paper presents a multimodal maritime dataset and the data collection procedure used to gather it, which aims to facilitate autonomous navigation in restricted water environments. The dataset comprises measurements obtained using various perception and navigation sensors, including a stereo camera, an infrared camera, an omnidirectional camera, three LiDARs, a marine radar, a global positioning system, and an attitude heading reference system. The data were collected along a 7.5-km-long route that includes a narrow canal, inner and outer ports, and near-coastal areas in Pohang, South Korea. The collection was conducted under diverse weather and visual conditions. The dataset and its detailed description are available for free download at https://sites.google.com/view/pohang-canal-dataset.

\end{abstract}


\maketitle

\section{Introduction}
Recent advances in autonomous vehicle technology have led to a growing interest in autonomous navigation in maritime environments. The development of advanced perception sensors with rapid progress of computer vision and deep learning have enabled researchers to explore the potential of autonomous surface vehicles (ASVs) in the maritime domain. However, unlike on-land applications such as self-driving cars, acquiring real-world datasets with sufficient quantity and variety for maritime environments is challenging. As a result, there is a shortage of such datasets available to researchers and developers in this field. 

Previous studies on maritime datasets have primarily focused on computer vision applications, specifically vessel detection and classification. While these visual datasets are useful for developing maritime computer vision algorithms, other types of perceptual and navigation data that are synchronously acquired onboard are necessary to perform multi-sensor fusion and achieve reliable and robust operation. For example, LiDARs and cameras may play a crucial role in tasks requiring precision, such as docking \citep{Docking2021}, object detection, and collision avoidance \citep{ARAGONUSV2020}, as well as autonomous navigation in narrow waterways \citep{Roboat2019}. Additionally, radar is capable of detecting objects in the distance \citep{RadarDetection2021} and can be used for vehicle localization through coastline detection \citep{CostalSLAM2019}. Therefore, it is essential to collect and analyze a diverse range of perceptual and navigation data to develop effective autonomous navigation systems for the maritime domain. 

\begin{figure}[tbh]
    \centering
    \includegraphics[width=0.98\linewidth]{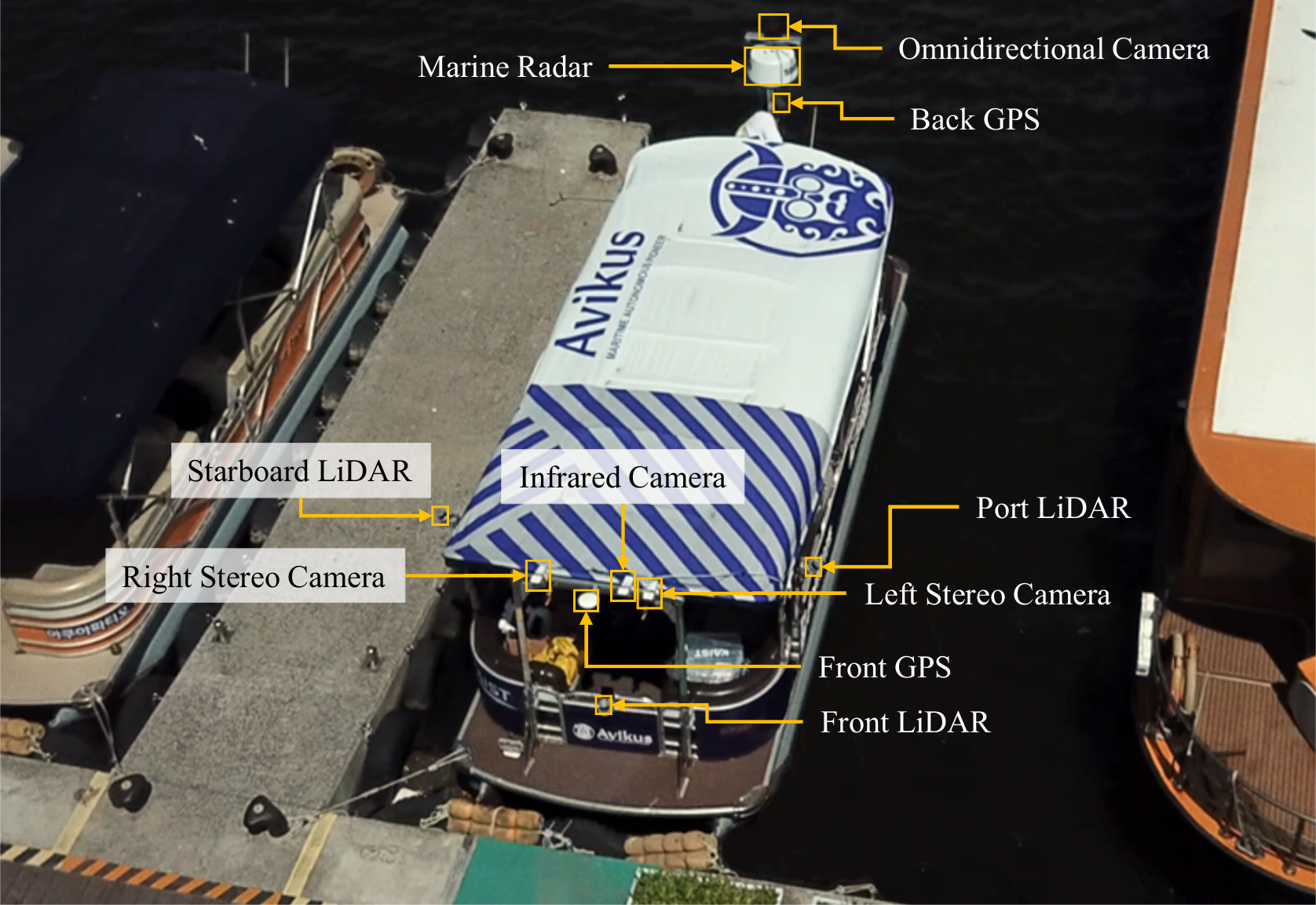}
    \caption{Vehicle used for data collection}
    \label{fig:platform}
\end{figure}

\begin{figure}[tbh]
    \centering
    \includegraphics[width=0.98\linewidth]{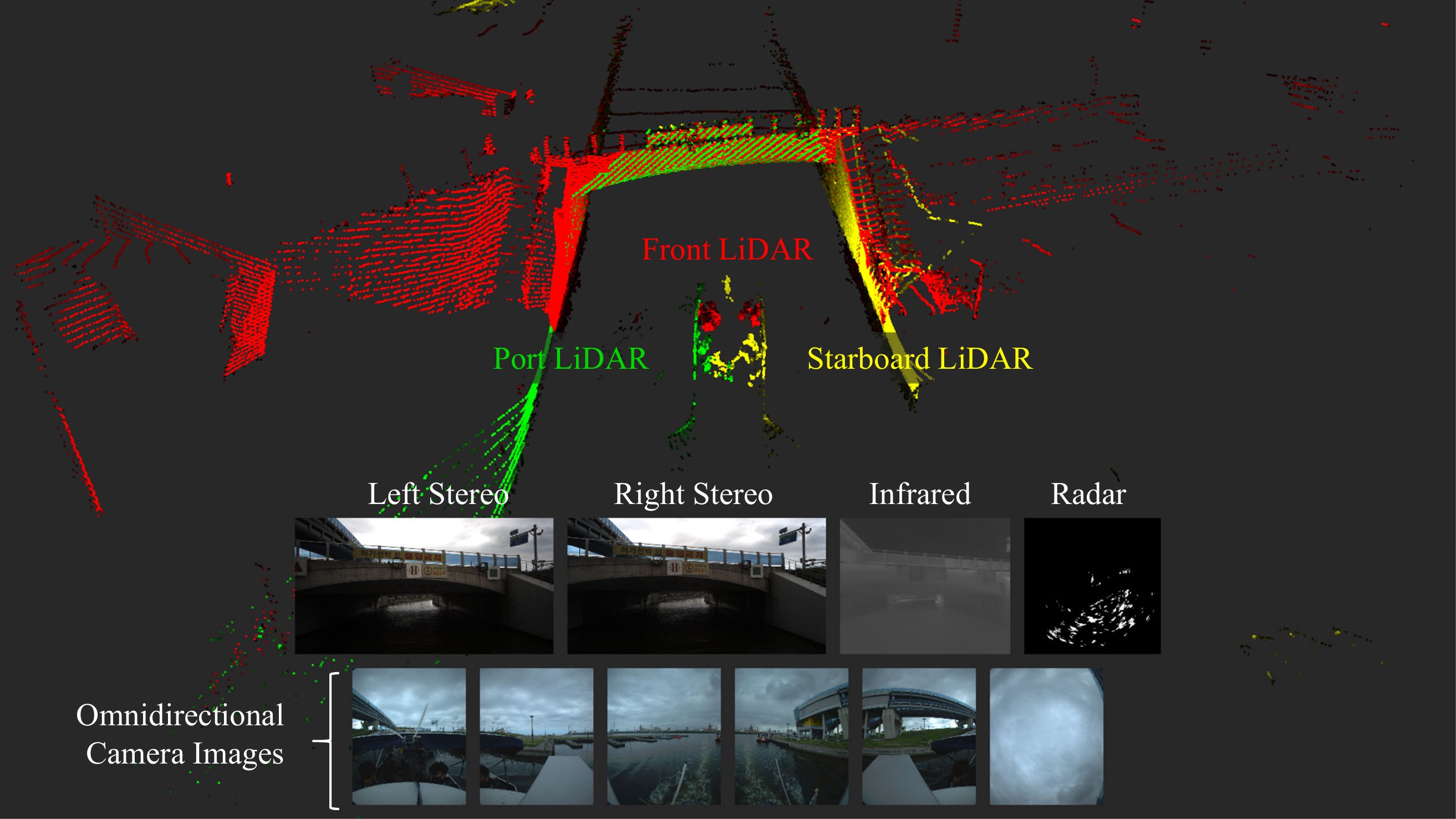}
    \caption{Example of the dataset}
    \label{fig:data_example}
\end{figure}

This paper presents a multimodal maritime dataset that includes diverse types of perceptual and navigation data to aid in developing autonomous navigation capabilities in restricted water environments such as narrow canals and port areas. The dataset was collected in Pohang, South Korea, following a 7.5-km-long route that comprised a narrow canal, port, and near-coastal areas. Figure \ref{fig:platform} shows the vehicle and sensor suite used for data acquisition, and Fig. \ref{fig:data_example} displays data from all of the perceptual sensors used in this study. The vehicle was equipped with two GPS antennas, one in the front and one in the back, as well as an attitude heading reference system (AHRS) mounted at the back of the vehicle. Two visual cameras that served as a stereo camera were mounted in the front of the vehicle, along with an infrared camera. Additionally, three LiDARs were mounted on the vehicle's front, port, and starboard sides. An omnidirectional camera and a marine radar were mounted on a vertical slide at the back of the vehicle, as shown in Figure \ref{fig:platform}. 

The remainder of this paper is organized as follows. In Section \ref{sec:related_work}, we summarize the existing open datasets related to maritime environments. Section \ref{sec:system_configuration} describes the system configuration, including the hardware sensor system and recording system. We present the sensor calibration methods used in this study in Section \ref{sec:sensor_calib}. In Section \ref{sec:canal_dataset}, we describe the configuration and characteristics of the dataset, including the environmental conditions and the structure of the dataset. Finally, the concluding remarks are provided in Section \ref{sec:conclusion}.

\section{Related work} 
\label{sec:related_work}

Real-world datasets are crucial for validating algorithms and learning materials in deep learning-based mobile robotics applications. Many on-land datasets have been published and used for evaluating performance, particularly in simultaneous localization and mapping (SLAM), as well as for data-intensive deep learning studies. These datasets include perceptual and navigation data collected in various environments, such as parks and campuses using unmanned ground vehicles \citep{NewCollege2009}, roads using cars \citep{KITTI2012}, and indoors using unmanned aerial vehicles (UAVs) \citep{EuRoC2016}. To enhance these datasets, researchers have explored different scenes \citep{ComplexUrban2019} and employed additional sensors like radars \citep{OxfordRadar2020, MulRAN2020, BOREAS2022}.

Previous research on maritime datasets has primarily focused on object detection and classification by providing image data with ground-truth annotations. For instance, in \cite{VAIS2015}, cropped visible and infrared images of vessels captured during both the day and night were provided for classification tasks. In \cite{ARGOS}, images obtained from surveillance cameras situated in buildings along the Grand Canal of Venice, Italy, were supplied with bounding box annotations. Furthermore, in \cite{Singapore2017}, visible and near-infrared videos captured from on-shore and onboard locations were provided with ground-truth annotations of horizon detection, object detection, and tracking data. In \cite{AirborneMaritime}, airborne surveillance images captured using a UAV were supplied with bounding box annotations for object detection and tracking. In \cite{MaSTr13252019}, onboard camera images with semantic segmentation labels of the sky, obstacles, and water were provided. In \cite{WaterSegment2019}, the authors focused on water segmentation through the visual images captured using an onboard camera. Finally, in \cite{LiDAR2022}, the authors presented simulated and real-world data of 3D LiDAR point clouds and ground-truth 3D bounding boxes, focusing on deep-learning-based object detection.

\begin{figure*}
     \centering
     \begin{subfigure}[b]{0.652\linewidth}
         \centering
    \includegraphics[width=1.0\linewidth]{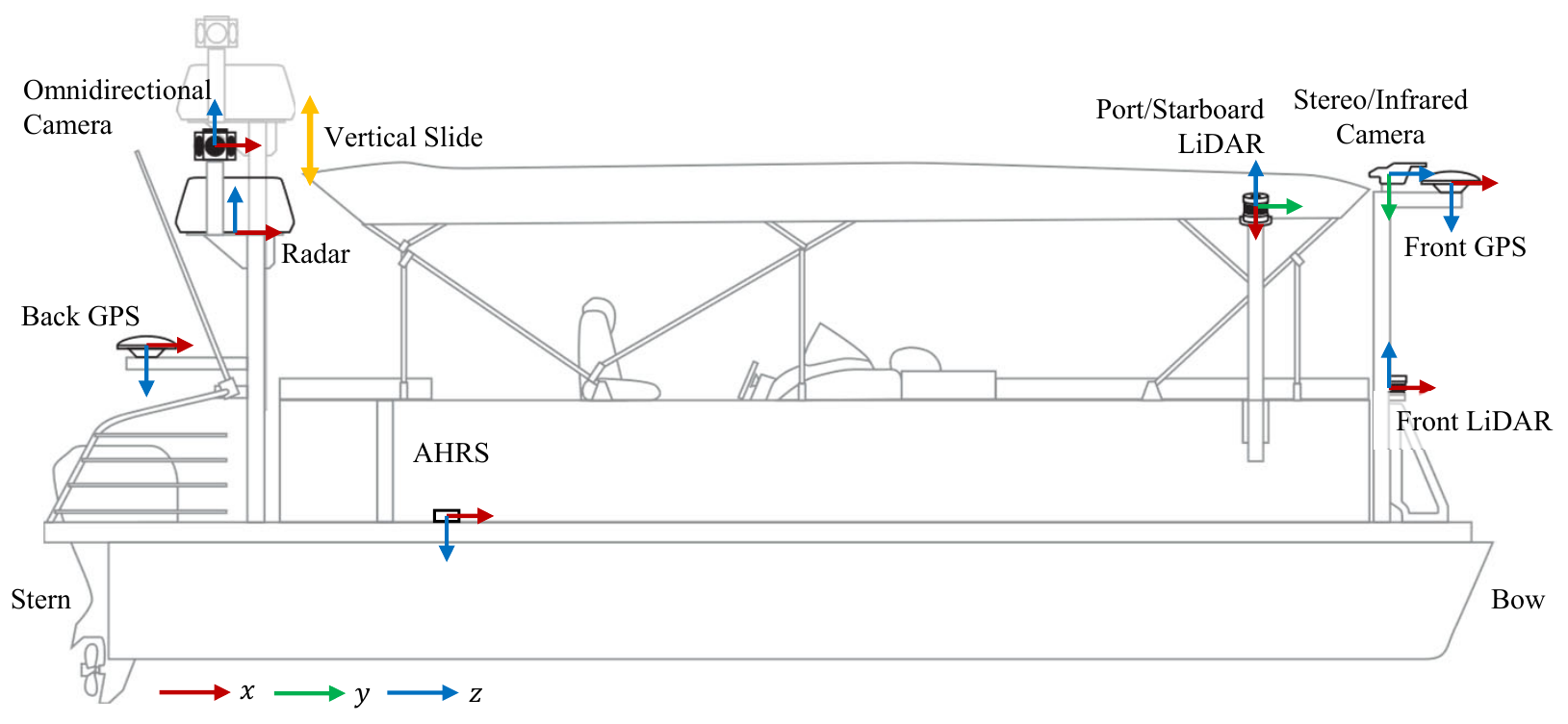}
         \caption{Side view}
         \label{fig:sensor_config_side}
     \end{subfigure}
     \hfill
     \begin{subfigure}[b]{0.302\linewidth}
         \centering
         \includegraphics[width=1.0\linewidth, trim={0 0 0 -0.2cm},clip]{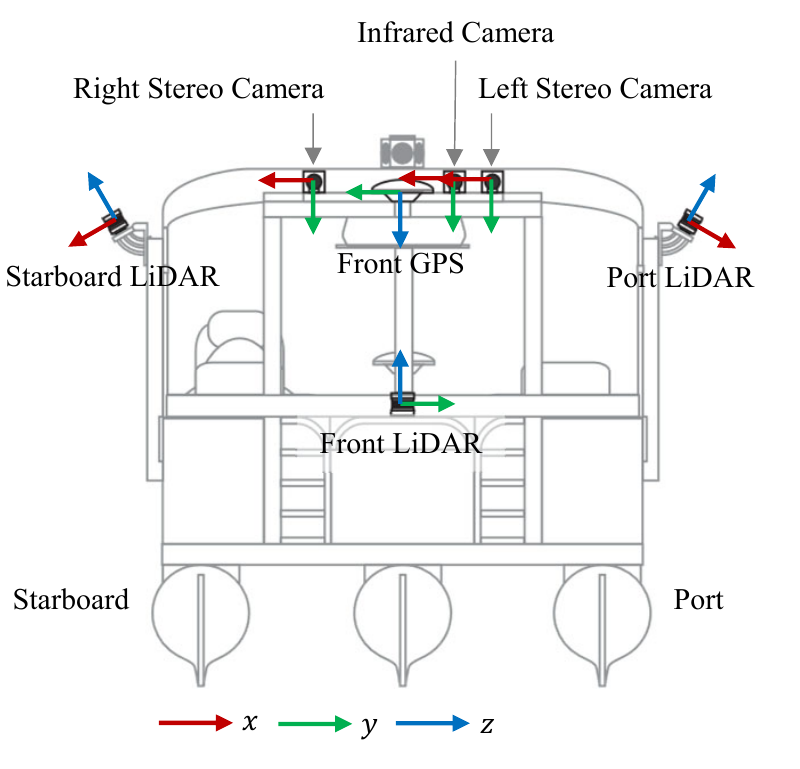}
         \caption{Front view}
         \label{fig:sensor_config_front}
     \end{subfigure}
     \hfill
     \begin{subfigure}[b]{0.53\linewidth}
         \centering
    \includegraphics[width=1.0\linewidth]{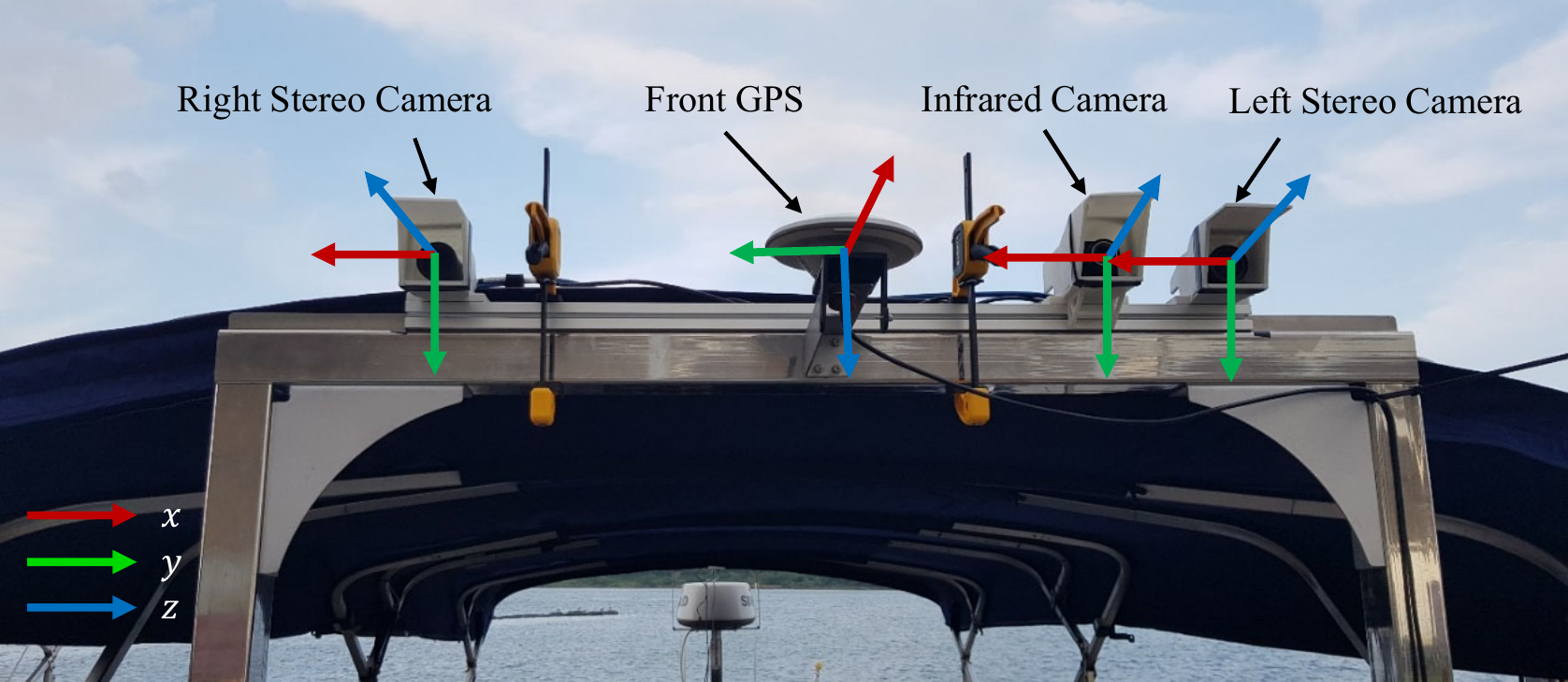}
         \caption{Camera module with two visible cameras and an infrared camera}
         \label{fig:camera_module}
     \end{subfigure}
     \hfill
     \begin{subfigure}[b]{0.44\linewidth}
         \centering
         \includegraphics[width=1.0\linewidth]{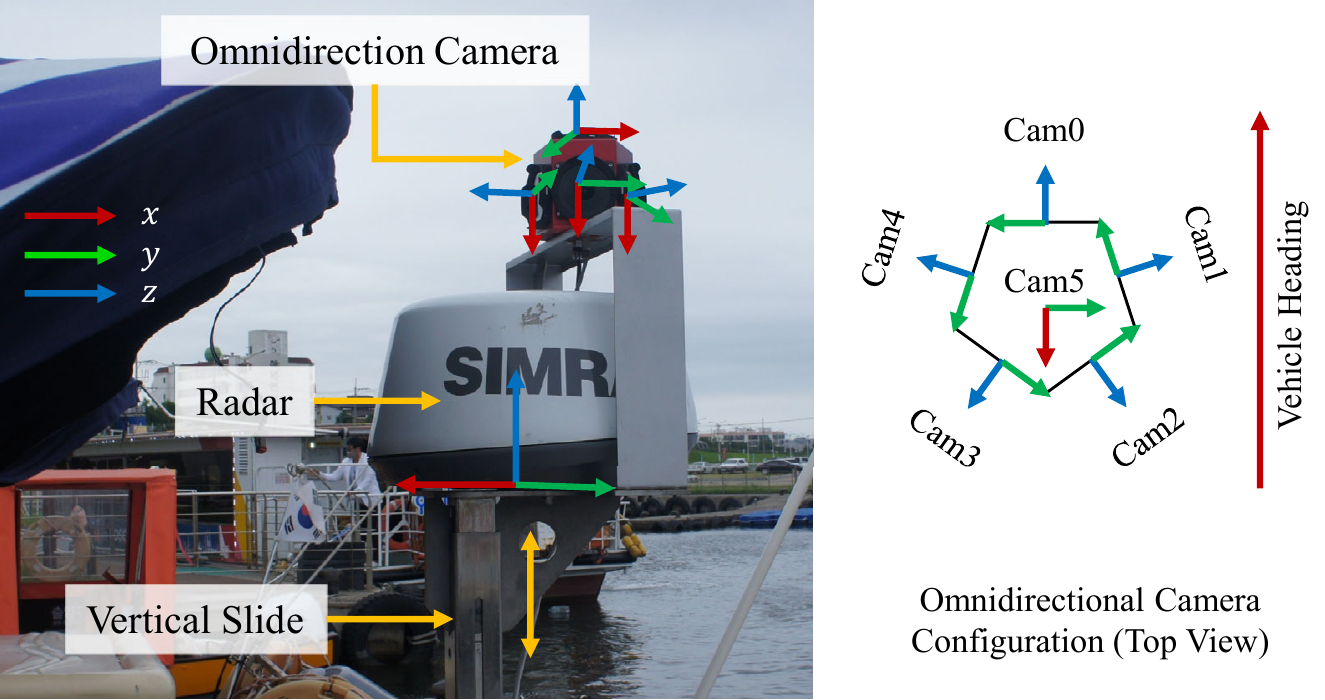}
         \caption{Omnidirectional camera and a marine radar}
         \label{fig:omni_radar_module}
     \end{subfigure}
     \hfill
    \caption{Hardware sensor configuration: a diagram of the vehicle viewed from the starboard side (a) and front side (b), a picture of the camera module (c), and the omnidirectional camera and radar mounted on the vertical slide (d). The coordinates are depicted using red (x), green (y), and blue (z) arrows.}
    \label{fig:sensor_config}
\end{figure*}

There is a scarcity of publicly available maritime datasets that contain both onboard perceptual and navigation data. Only a few datasets such as \cite{MODD22018} and \cite{USVInland} have been made available. The former provides stereo camera images, GPS, and inertial measurement unit (IMU) data for semantic segmentation of waterways, while the latter presents onboard perceptual and navigation data collected in different weather conditions in an inland waterway using a small unmanned surface vehicle. This dataset includes a stereo camera, a 16-channel 3D LiDAR, three millimeter-wave radars, a GPS, and an onboard IMU. It also provides ground-truth annotations of water segmentation and the ground-truth trajectory using GPS and IMU data.

\section{System configuration}
\label{sec:system_configuration}

\subsection{Sensor configuration}
\label{subsec:sensor_configuration}

The sensor configuration of the vehicle is shown in Figure \ref{fig:sensor_config}. The vehicle used in the study was a 7.9 m long and 2.6 m wide cruise boat, with a weight of 1.7 tons and a capacity of 12 persons. Two GPS antennas were mounted on the front and back of the vehicle, respectively, with a baseline of 6.8 m. They were connected to a global navigation satellite system (GNSS) with real-time kinematic (RTK) receiver, providing location with an RTK accuracy of 0.01 m + 1 ppm and heading measurement with an error smaller than 0.28$\degree$. The GPS data was recorded at a frequency of 5 Hz. An attitude and heading reference system (AHRS) was placed on the floor at the back of the vehicle. The AHRS provided acceleration measurements with a resolution of 0.02 mg and bias of 0.04 mg, gyroscopic measurements with a resolution of 0.003$\degree/s$ and bias of 8$\degree/h$, and attitude measurements with a root mean square error of 0.5$\degree$ along the pitch and roll directions. The magnetometer of the AHRS was calibrated using the provided software. The AHRS data was recorded at a frequency of 100 Hz.

To enhance the detection of nearby objects and safety hazards, three LiDARs were installed on the front and sides of the vehicle. A 64-channel LiDAR was mounted horizontally at the front to detect obstacles or vehicles along the path. Two 32-channel LiDARs were positioned on the port and starboard sides of the vehicle to extend the measurement coverage. These side LiDARs were installed at a downward heading of 30$\degree$ relative to the horizontal plane to detect small objects on the surface that may pose a threat to vehicle operation. Each LiDAR had a range of 120 m, with a resolution of 0.3 cm and a precision of 1.0 to 5.0 cm. The horizontal field of view (FOV) was 360$\degree$ with a resolution of 2048, and the vertical FOV of all LiDARs was 33.2$\degree$. The vertical resolution varied depending on the number of channels (64 for the front LiDAR and 32 for the port and starboard LiDARs). The point cloud data collected from each LiDAR were gathered asynchronously at 10 Hz per cycle. A frequency-modulated continuous-wave (FMCW) radar with an X-band wavelength (9.3 GHz to 9.4 GHz) was mounted at the back of the vehicle. The radar had a range of 50 to 1654.8 m and rotated 1.0 to 1.3 times per second.

The camera module, as depicted in Fig.~\ref{fig:camera_module}, was mounted at the front of the vehicle and equipped with two visual cameras and an infrared camera, all with 3D-printed lens hoods. The two visual cameras were configured as a stereo camera and synchronized at the hardware level using a trigger cable. Images were acquired simultaneously from both cameras at a frequency of 10 Hz. The infrared camera had a temperature measurement range of -25 to 135°C, and automatic thermal calibration at the hardware level was performed during data collection to compensate for device temperature variations. Thermal images were acquired at a rate of 10 Hz. An omnidirectional camera and a marine radar were installed on a vertical slide at the back of the vehicle. The slide was kept at its lowest position while the vehicle passed under low bridges and was manually raised to obtain better views. The omnidirectional camera captured six images (five horizontal views and one top view) simultaneously at a frequency of 10 Hz. The models and detailed specifications of all sensors are available on our dataset website.

\subsection{Data collection system}
\begin{figure}[tbh]
    \centering
    \includegraphics[width=0.95\linewidth, trim={0 0 0 -0.3cm},clip]{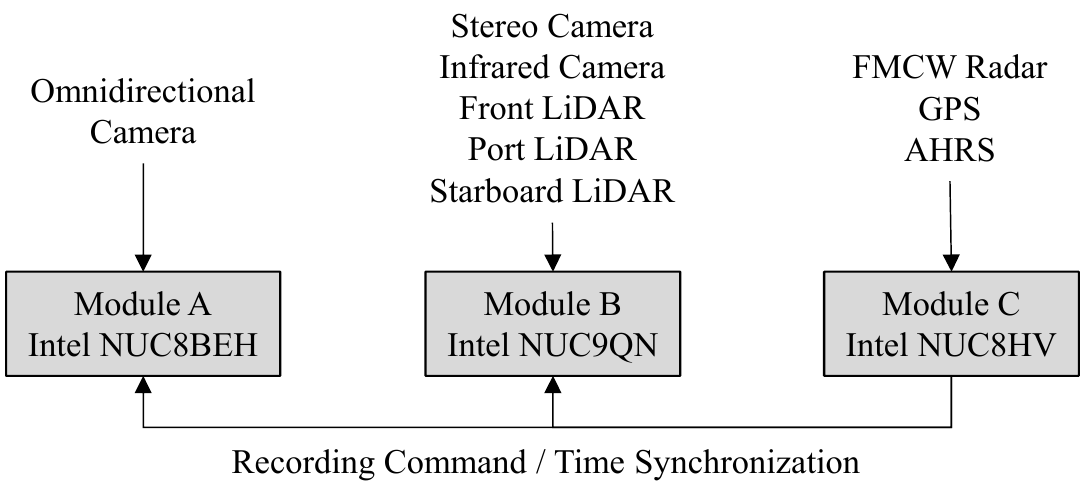}
    \caption{Data collection system}
    \label{fig:data_collection_system}
\end{figure}

Figure \ref{fig:data_collection_system} shows a schematic diagram of the data collection process, which involved three computers, referred to as modules A, B, and C. Module A recorded the images captured using the omnidirectional camera, module B recorded left and right stereo images, infrared images, and point clouds of the front, port, and starboard LiDARs, while module C recorded radar images, GPS signals, and IMU data. The system times of the three computers were synchronized tightly using the Chrony \citep{Chrony} library, and the system times of modules A and B were periodically synchronized with module C. The data recording initiation and termination were carried out by sending commands from module C to modules A and B.

In order to avoid delays and loss of data during the simultaneous streaming of data from multiple sensors, the recording programs for each module were designed to be more efficient. To address the issue in module A, where converting the omnidirectional camera image stream to image files was slowing down the recording process, the image streams were saved as binary files and converted during post-processing. In module B, three individual processes were run simultaneously to record stereo image pairs, infrared images, and data from the three LiDARs. Multiple threads within each process were activated to prevent data loss. On the other hand, module C required only a single-threaded process for each sensor measurement, as the data size for each sensor was relatively small.

\section{Sensor calibration}
\label{sec:sensor_calib}

\subsection{Intrinsic calibration of cameras}
For accurate estimation of the intrinsic parameters of the cameras, two structured calibration boards were used for the visual and infrared cameras. The intrinsic calibration of each camera was performed using the method described in \cite{CalibrationZhang2000}, which involves using a checkerboard pattern calibration board for the visual cameras in the stereo camera and omnidirectional camera, and a custom-made calibration board that can reflect infrared lights for the infrared camera.

\subsection{Extrinsic calibration}

Extrinsic calibration between the perceptual sensors was performed using either the calibration board or unstructured measurement data, taking into account the overlap between their measurement ranges and the limitations of the experimental environment. In the case of the stereo camera, the images had sufficient overlapping regions, allowing for most of the co-visible area to be covered using the calibration board. However, since the omnidirectional camera was located at the back of the vehicle, several portions of the images were occluded by the vehicle, resulting in an inadequate co-visible area for reliable calibration.

The extrinsic parameters between the front LiDAR and other perceptual sensors were estimated using an in-motion calibration method, due to the vehicle's movement and the asynchronous sensor measurements. This method involved constructing a point cloud map using a LiDAR SLAM approach and estimating the LiDAR poses from the SLAM result. Next, the sensor pose in the global coordinate system was estimated by matching the sensor data to the point cloud map (refer to sections \ref{sec:lidarToLidarCalib} to \ref{sec:lidarToCameraCalib}). The LiDAR pose when the sensor data was collected was determined by interpolating the given LiDAR poses using the measurement timestamps. Finally, the extrinsic parameters were calculated using the estimated LiDAR pose and the sensor pose in the global coordinate system.

\begin{figure}
    \centering
    \begin{subfigure}[t]{0.92\linewidth}
        \centering
        \includegraphics[width=1.0\linewidth, trim={0 0 0 -0.3cm},clip]{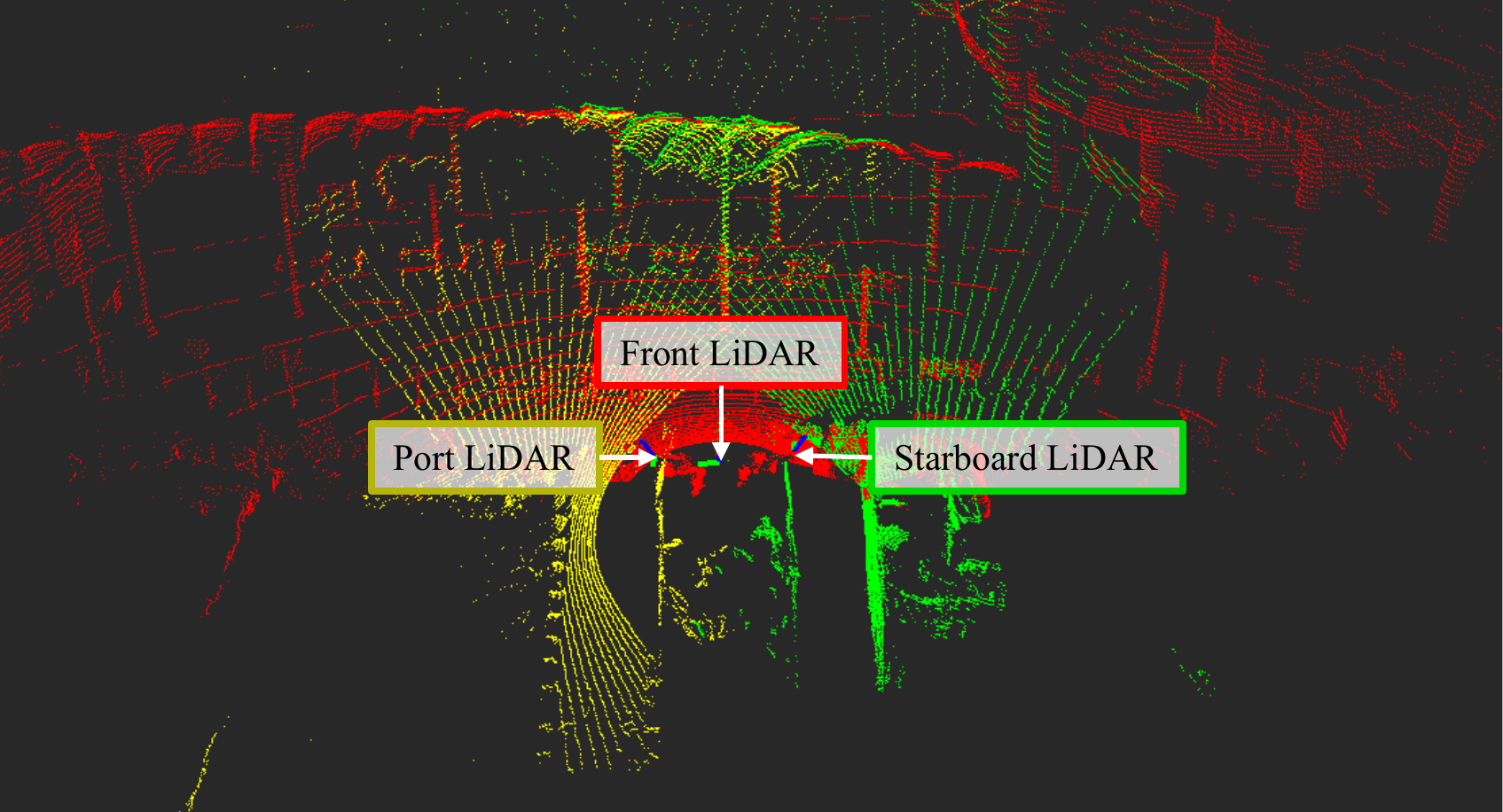}
        \caption{Point clouds obtained from the front (red), port (yellow), and starboard (green) LiDARs}
        \label{fig:calib_lidar_lidar}
    \end{subfigure}
    \hfill
    \begin{subfigure}[t]{0.92\linewidth}
        \centering
        \includegraphics[width=1.0\linewidth]{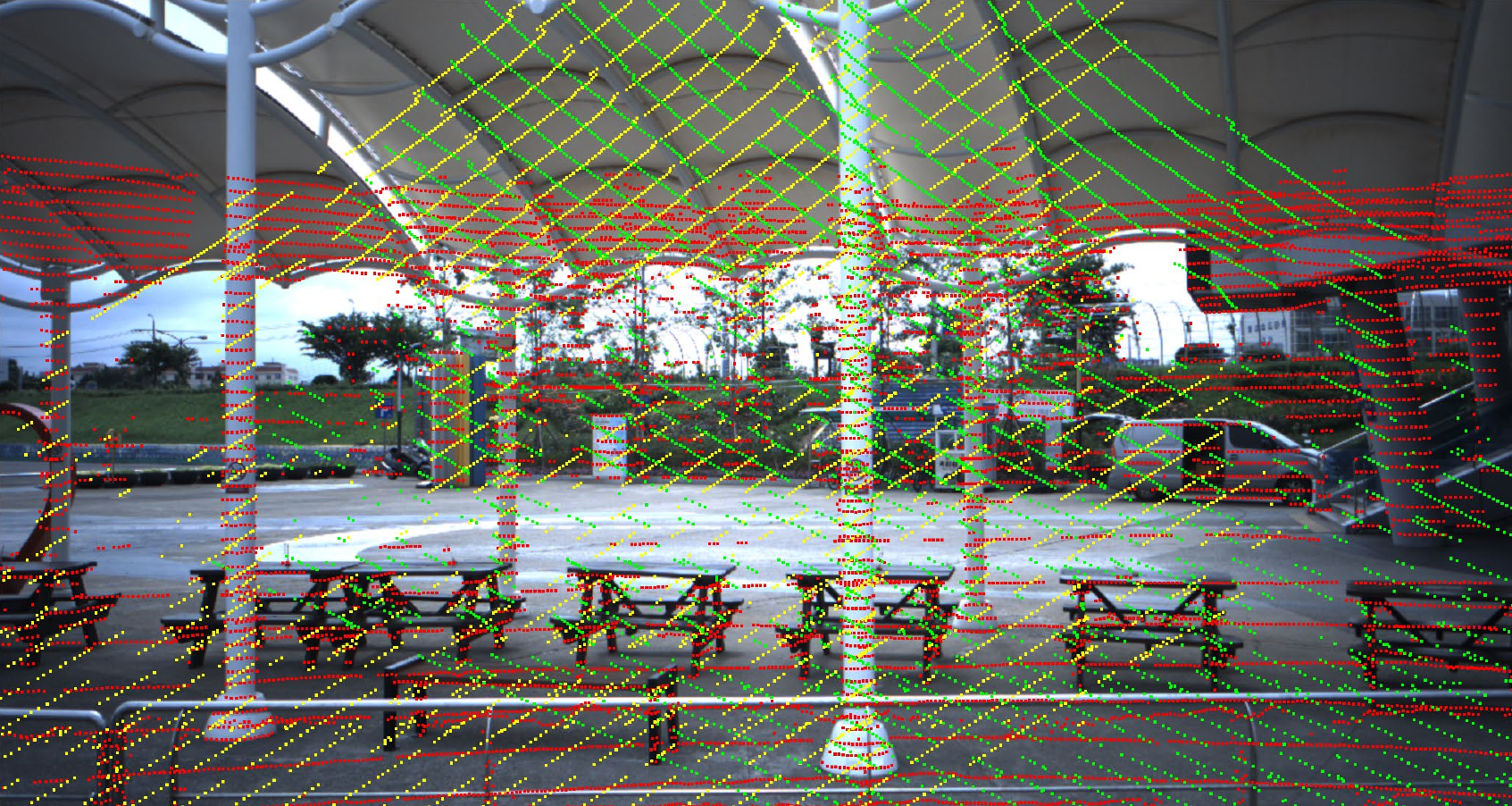}
        \caption{Projection of point cloud onto the left stereo camera}
    \label{fig:calib_lidar_camera}
    \end{subfigure}
    \hfill
    \begin{subfigure}[t]{0.92\linewidth}
        \centering
        \includegraphics[width=1.0\linewidth]{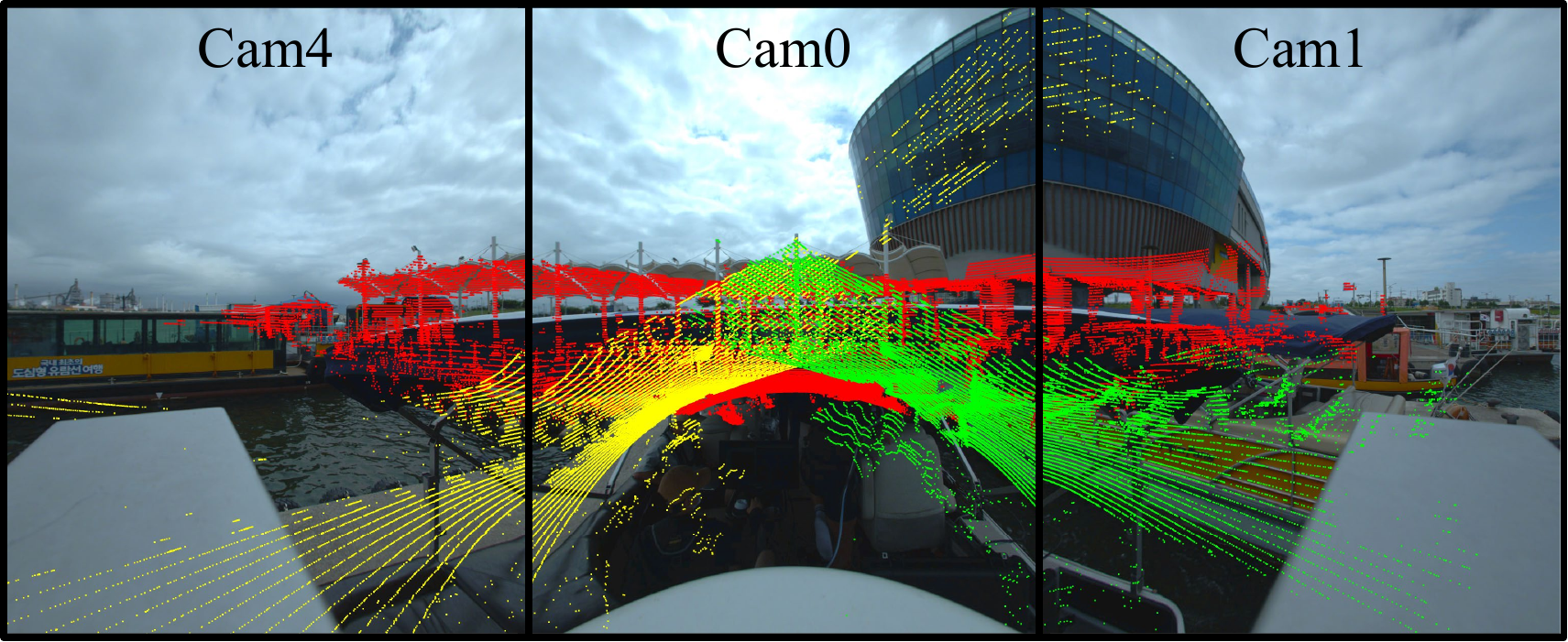}
        \caption{Projection of point cloud onto the omnidirectional camera}
    \label{fig:calib_lidar_omni}
    \end{subfigure}
    \hfill
    \caption{Example of LiDAR data and their projection onto the cameras}
    \label{fig:calib_lidar_camera_projection}
\end{figure}

\subsubsection{Omnidirectional camera calibration.}

The extrinsic calibration for the cameras in the omnidirectional camera was achieved by utilizing ArUco markers \citep{ARUCO2018} to improve the correspondence detection between images, as the overlapping regions between them were narrow. A factor graph framework was used with camera pose nodes and 3D point nodes of the calibration board in GTSAM \citep{GTSAM} to achieve accurate calibration.

\subsubsection{LiDAR to LiDAR calibration.}
\label{sec:lidarToLidarCalib}

The point clouds obtained from each LiDAR are shown in Figure \ref{fig:calib_lidar_lidar}. The initial calibration was performed by manually adjusting the rotation and translation of the port and starboard LiDARs. Then, to achieve fine calibration between the front/port LiDARs and front/starboard LiDARs, the Generalized Iterative Closest Points (GICP) method \citep{GICP2009} was applied to each port and starboard point cloud to align them with the point cloud map generated by the front LiDAR.

\subsubsection{LiDAR to camera calibration.}
\label{sec:lidarToCameraCalib}

The extrinsic calibration between the front LiDAR and cameras was performed by solving linear PnP (perspective-n-points) \citep{PnP1999} problems using carefully selected points in the images and corresponding points in the point cloud map generated with the front LiDAR. Figure~\ref{fig:calib_lidar_camera} shows the calibration results for the LiDARs to the left stereo camera, while Figure~\ref{fig:calib_lidar_omni} shows the results for the omnidirectional camera.

\subsection{Hand-eye calibration}
As there is no perceptual correlation between LiDAR to GPS and AHRS, we utilized the hand-eye calibration method \citep{HandEye1989} for extrinsic calibration between AHRS to LiDAR and GPS.

\subsubsection{AHRS to LiDAR calibration.}
The positional drift due to accelerometer error integration can be severe. However, the orientation can be precisely measured, as the AHRS provides attitude measurements from the direction of gravity and geomagnetic force. Due to these limitations, only the rotation component of the AHRS to LiDAR calibration is obtained using the hand-eye calibration method, and the relative translation is determined based on the ASV blueprint. The dataset contains large rotational movements near a feature-rich pier area, so orientation measurements from the AHRS and LiDAR odometry were sampled to estimate the relative rotation by solving the hand-eye problem.

\subsubsection{AHRS to GPS calibration.}
\begin{figure}[tbh]
    \centering
    \includegraphics[width=0.7\linewidth]{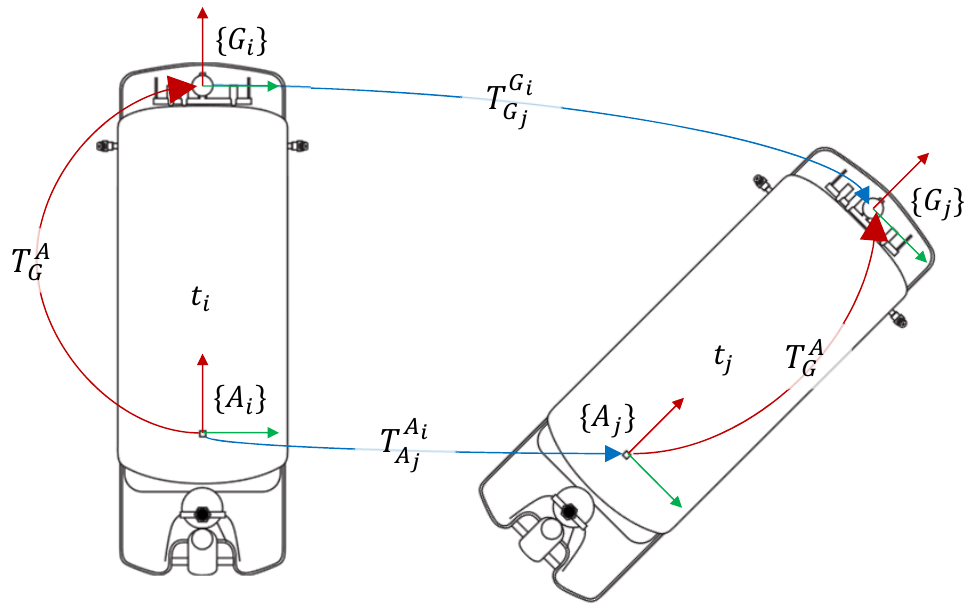}
    \caption{Top view of the vehicle at time $t_i$ and $t_j$. The AHRS and GPS coordinate systems are shown as \{A\} and \{G\}.}
    \label{fig:hand_eye_calib}
\end{figure}

\begin{figure}[tbh]
    \centering
    \includegraphics[width=0.98\linewidth]{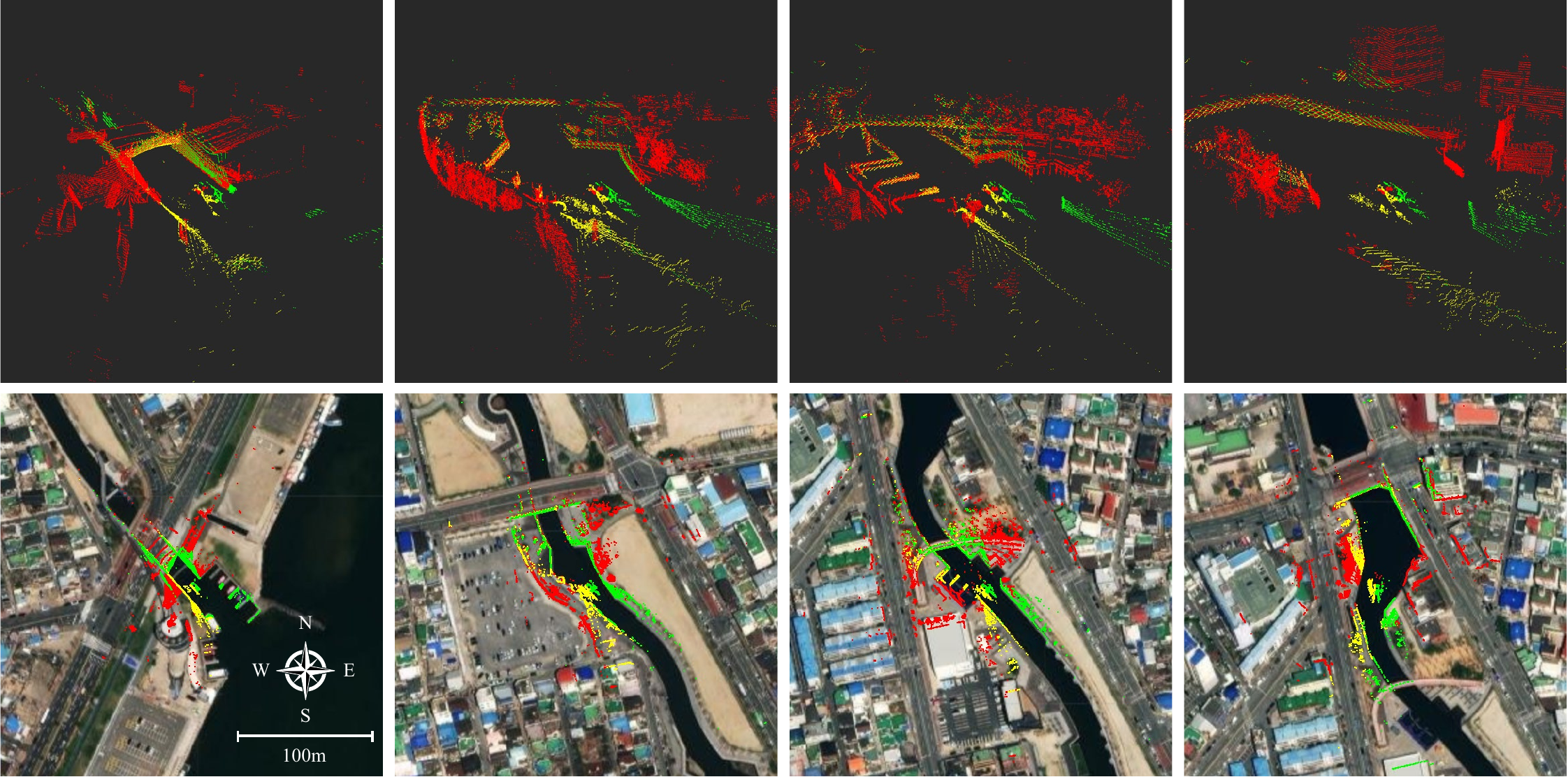}
    \caption{Point clouds of LiDARs and their projections on the satellite map based on GPS data}
    \label{fig:calib_lidar_gps}
\end{figure}

The extrinsic calibration between the AHRS and GPS was performed using both the LiDAR-inertial odometry and GPS 2D pose measurements. As the GNSS-RTK provided accurate location and heading information, both translation and rotation were considered for the relative pose estimation. Near the pier, where roll and pitch motions were negligible, we sampled data and used the 2D poses from the LiDAR-inertial odometry and GPS measurements to calculate translation matrix from AHRS to GPS, $T_G^A$. This equation relates the transformation matrix between the AHRS coordinate systems at time $t_i$ and $t_j$, $T_{A_j}^{A_i}$, and the transformation matrix between GPS coordinate systems at time $t_i$ and $t_j$, $T_{G_j}^{G_i}$, as shown in Fig.~\ref{fig:hand_eye_calib}. Figure \ref{fig:calib_lidar_gps} shows the extrinsic calibration result, where the LiDAR point clouds were projected onto satellite maps based on GPS measurements.

\subsection{Baseline trajectory}
\begin{figure}[tbh]
    \centering
    \includegraphics[width=0.95\linewidth]{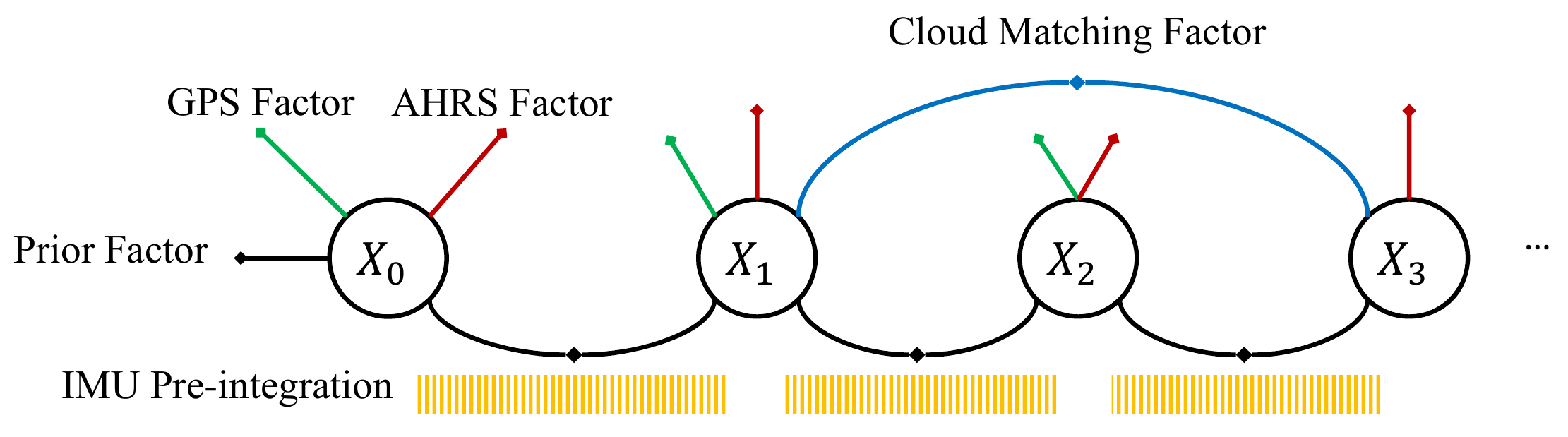}
    \caption{Factor graph model for baseline trajectory estimation}
    \label{fig:baseline_graph}
\end{figure}

While the GNSS-RTK delivers highly accurate pose information in most areas, it is constrained in regions where the vehicle passes under bridges. Therefore, a straightforward GNSS-INS navigation system may not be dependable in these areas. To address this issue, we utilized the front LiDAR measurements in conjunction with the GNSS-RTK and AHRS measurements in an incremental smoothing and mapping (iSAM) \citep{Kaess2008} graph SLAM framework that was implemented in GTSAM. The graph structure for the SLAM framework is depicted in Figure \ref{fig:baseline_graph}.

The LiDAR-inertial odometry framework utilized in this study is based on LIO-SAM \citep{LIOSAM2020}, with modifications made to the point cloud feature extraction method. Unlike urban environments where many objects can be detected as dense point clouds, the objects in the canal area are mainly detected as sparse point clouds. Additionally, the water surface in maritime environments does not provide sufficient information for LiDAR odometry performance, as the water absorbs the light. Therefore, a reliable feature extraction method is crucial for accurate LiDAR odometry in maritime environments.

\begin{figure}
    \centering
    \begin{subfigure}[t]{0.98\linewidth}
        \centering
        \includegraphics[width=1.0\linewidth]{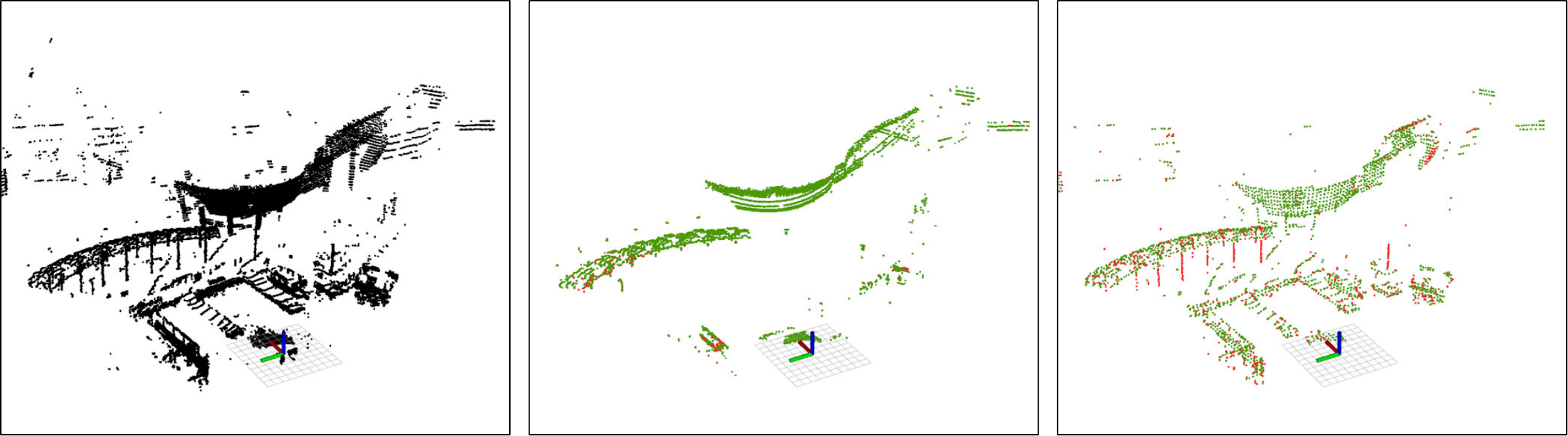}
        \caption{Pier}
        \label{fig:map_and_trajectory}
    \end{subfigure}
    \begin{subfigure}[t]{0.98\linewidth}
        \centering
        \includegraphics[width=1.0\linewidth]{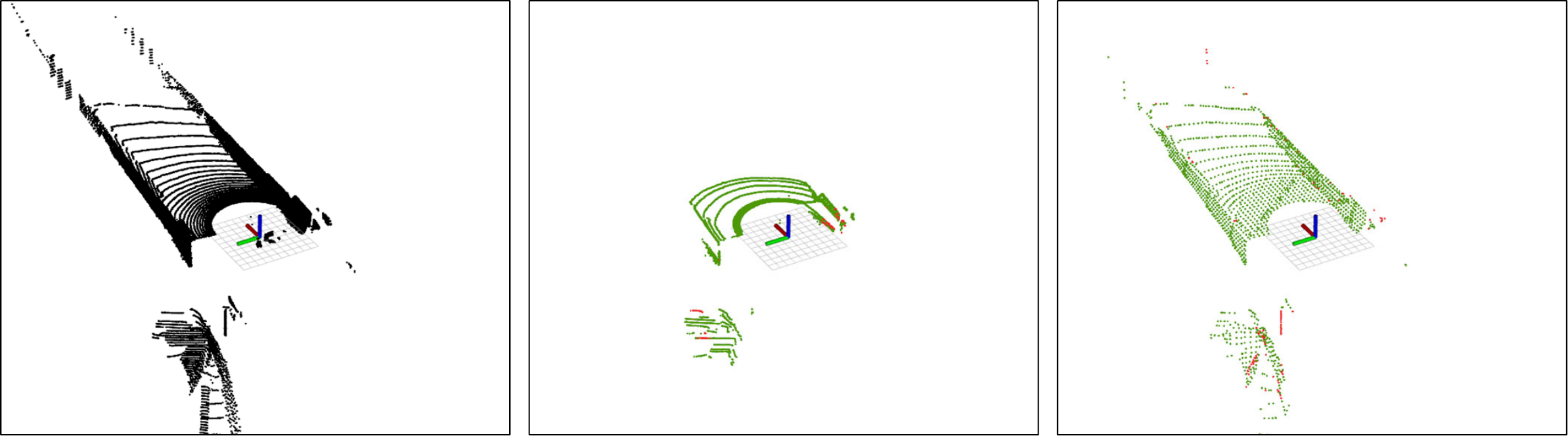}
        \caption{Under the bridge}
        \label{fig:map_lidar_radar}
    \end{subfigure}
    \caption{Comparison of feature extraction results in different places. The images on the left display the original point cloud, the middle images depict the feature extraction outcomes using LIO-SAM, and the right images represent the modified feature extraction results. The green points indicate surface points, and the red points indicate corner points.}
    \label{fig:lidarFeatures}
\end{figure}

\begin{figure}[tbh]
    \centering
    \includegraphics[width=0.98\linewidth]{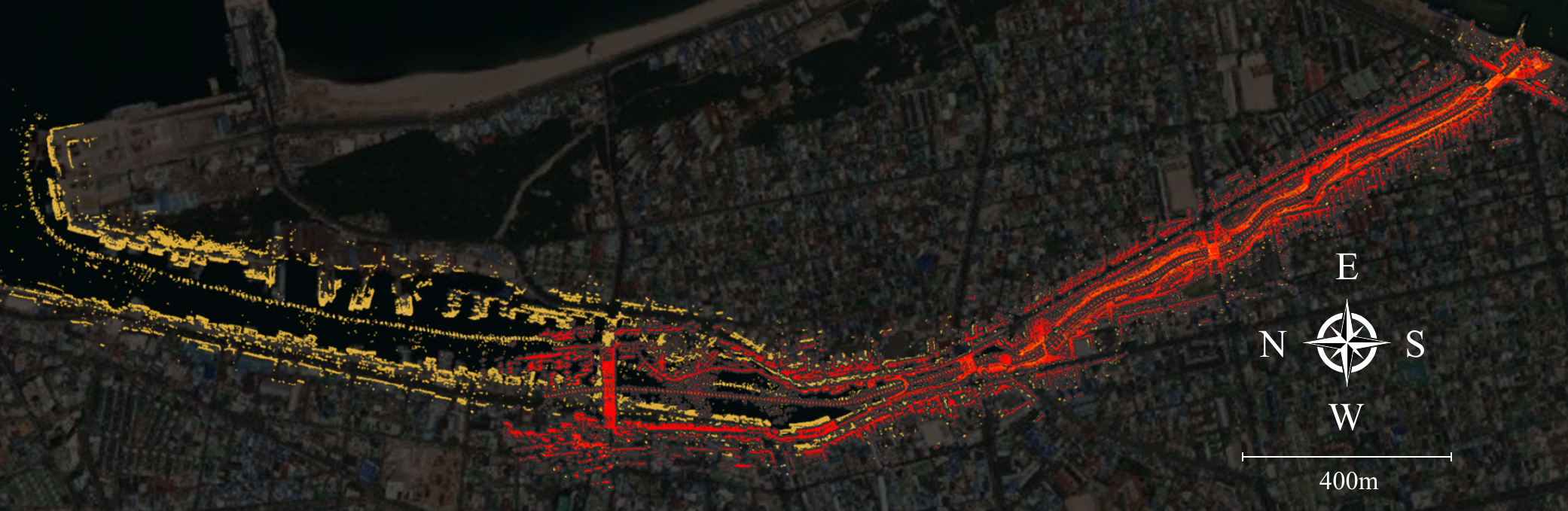}
    \caption{LiDAR-inertial SLAM result using LIO-SAM (red point cloud) and modified method (yellow) in the canal area to inner-port area.}
    \label{fig:lidarSlam}
\end{figure}

The feature extraction method in LIO-SAM, which is based on LOAM \citep{Zhang2014}, determines whether a point is a corner point or a surface point based on the curvature of the point and its neighboring points. However, this method often rejects points that have few neighboring points or are occluded, which can result in the rejection of points at a distance and thin pole-like objects. To improve the performance of LiDAR odometry in the canal area, we modified the point rejection algorithm by skipping occluded neighboring points and defining pole-like objects as corner features. Figure \ref{fig:lidarFeatures} illustrates the point cloud and feature extraction results of LIO-SAM and the modified algorithm. Figure \ref{fig:lidarSlam} shows the point cloud mapping results from the canal area to the inner-port area using LIO-SAM and the modified method, overlaid on the satellite map.

The GPS factor is created by inserting UTM coordinates and heading measurements obtained from the GPS, while the AHRS factor is created by inserting the orientation measurements obtained from the AHRS. However, since the direction of the true north measured by the GPS is different from the magnetic north measured by the AHRS, we only used the gravity direction estimated from the AHRS orientation measurement to correct the roll and pitch of the vehicle.

\section{The canal dataset}
\label{sec:canal_dataset}

\begin{figure*}
    \centering
    \begin{subfigure}[t]{0.98\linewidth}
        \centering
        \includegraphics[width=1.0\linewidth]{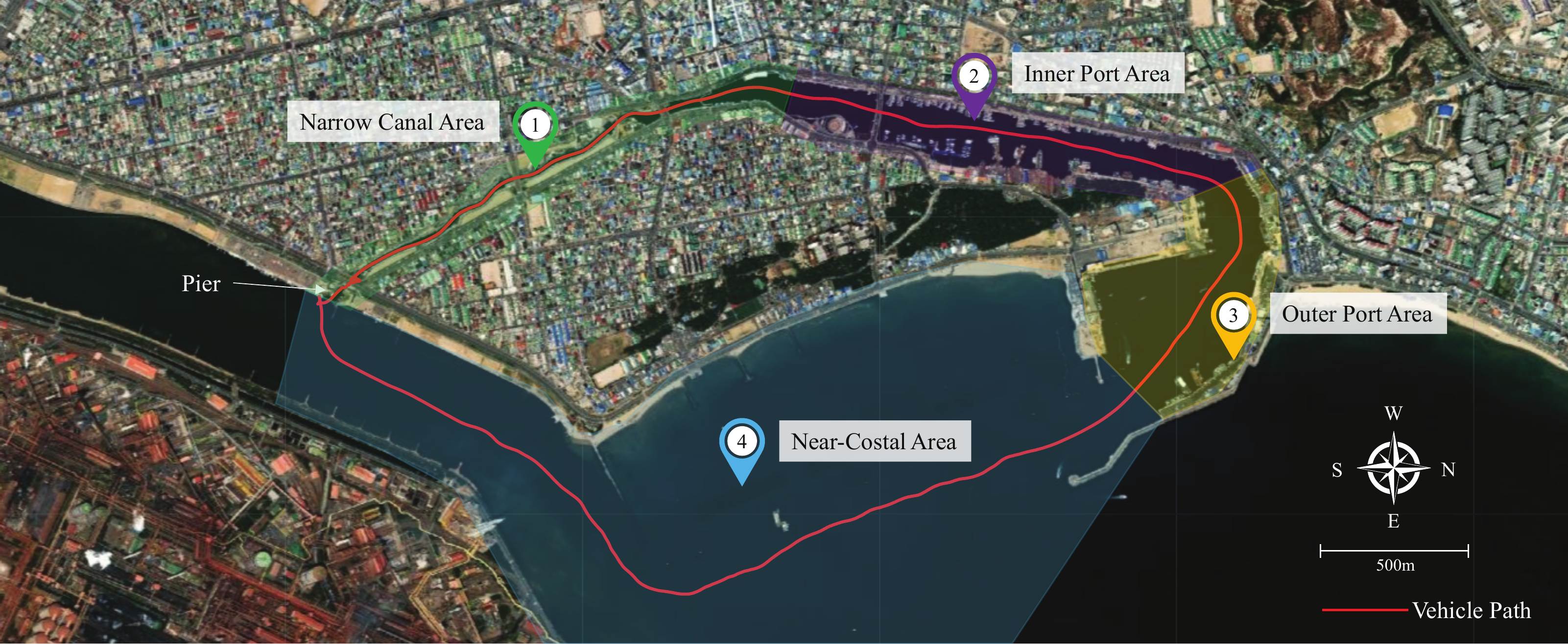}
        \caption{Map and vehicle trajectory}
        \label{fig:map_and_trajectory}
    \end{subfigure}
    \begin{subfigure}[t]{0.98\linewidth}
        \centering
        \includegraphics[width=1.0\linewidth]{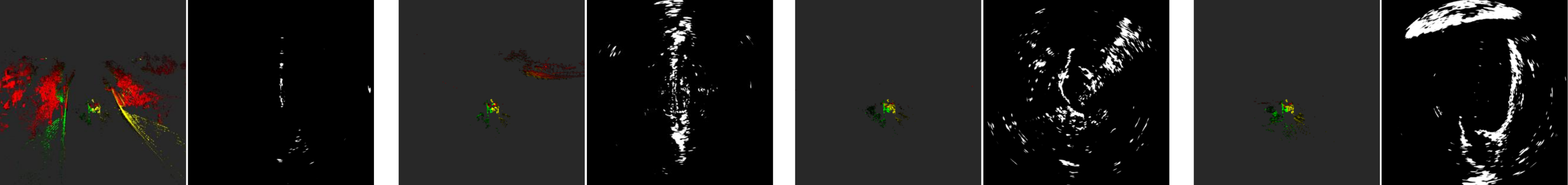}

        \caption{LiDAR and radar data}
        \label{fig:map_lidar_radar}
    \end{subfigure}
    \begin{subfigure}[t]{0.98\linewidth}
        \centering
        \includegraphics[width=1.0\linewidth]{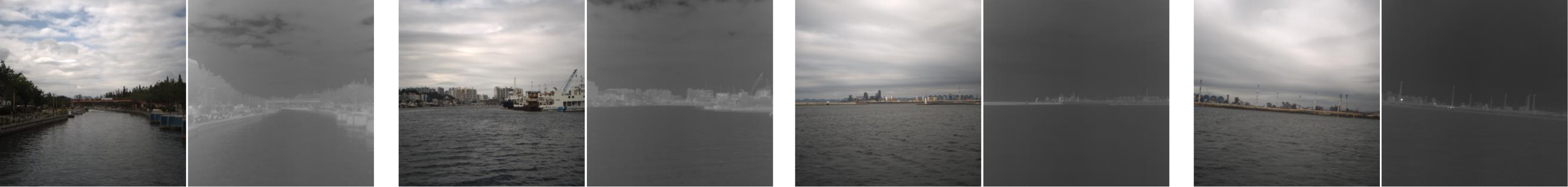}
        \caption{Visible and infrared images during daytime}
        \label{fig:map_EO_IR_day}
    \end{subfigure}
    \begin{subfigure}[t]{0.98\linewidth}
        \centering
        \includegraphics[width=1.0\linewidth]{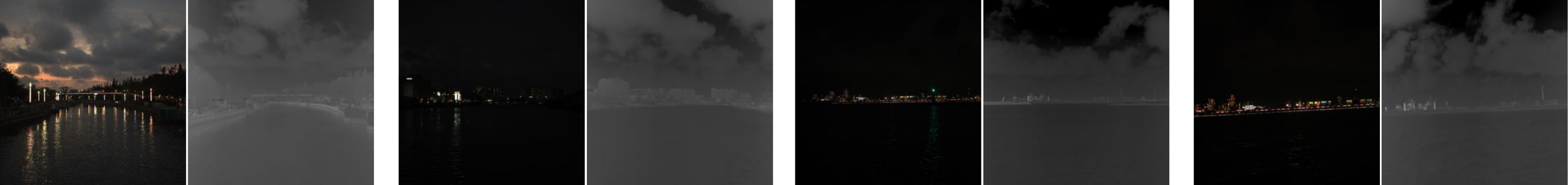}
        \caption{Visible and infrared images during nighttime}
        \label{fig:map_EO_IR_night}
    \end{subfigure}
    \caption{Data acquisition map, trajectory, and examples of data}
    \label{fig:map_and_scene}
\end{figure*}

The data used in the study was gathered from Pohang Canal, located in South Korea. The canal area consists of several regions, including a narrow canal area, an inner port area, an outer port area, and a near-coastal area, with a total length of 7.5 km. Figure \ref{fig:map_and_scene} shows the map of the experimental location, the trajectory of the vehicle, and examples of the collected data. The images in columns (b), (c), and (d) correspond to locations 1, 2, 3, and 4, respectively, as depicted in Fig.~\ref{fig:map_and_scene}.

\subsection{Experimental environment}
The main canal area is approximately 15 to 30 m wide and surrounded by a park and small buildings. In this region, LiDAR data produced a dense point cloud with rich geometrical features, while radar data was mostly obstructed by buildings and contained noise. The inner port area was split into two sides; on the west, there were fish markets and shops, while on the east, there were shipyards. Consequently, numerous fishing boats docked on the west side, while ships of various sizes were located on the east side. The water width in this area ranged from 50 to 100 m, and LiDARs acquired a sparse point cloud in the distance. Seawalls, passenger terminals, and storage buildings were present in the outer port area, alongside large cruise ships, cargo ships, and coast guard ships. LiDAR data hardly captured point clouds of on-land objects in this region, while radar data clearly revealed the coastline in the distance. A vast steelworks plant was located southeast of the experimental site, and it was mostly visible from the near-coastal area. The vehicle experienced relatively significant roll and pitch movements in the near-coastal area due to a sailing speed of around 10 knots, while the vehicle speed was 5 to 7 knots in the canal and port areas. The omnidirectional camera was removed during nighttime data collection due to safety concerns. Detailed explanations and characteristics of each dataset are available on our dataset website.

\subsection{Dataset structure}
\label{subsec:data_structure}

\begin{figure*}[tbh]
    \centering
    \includegraphics[width=0.98\linewidth]{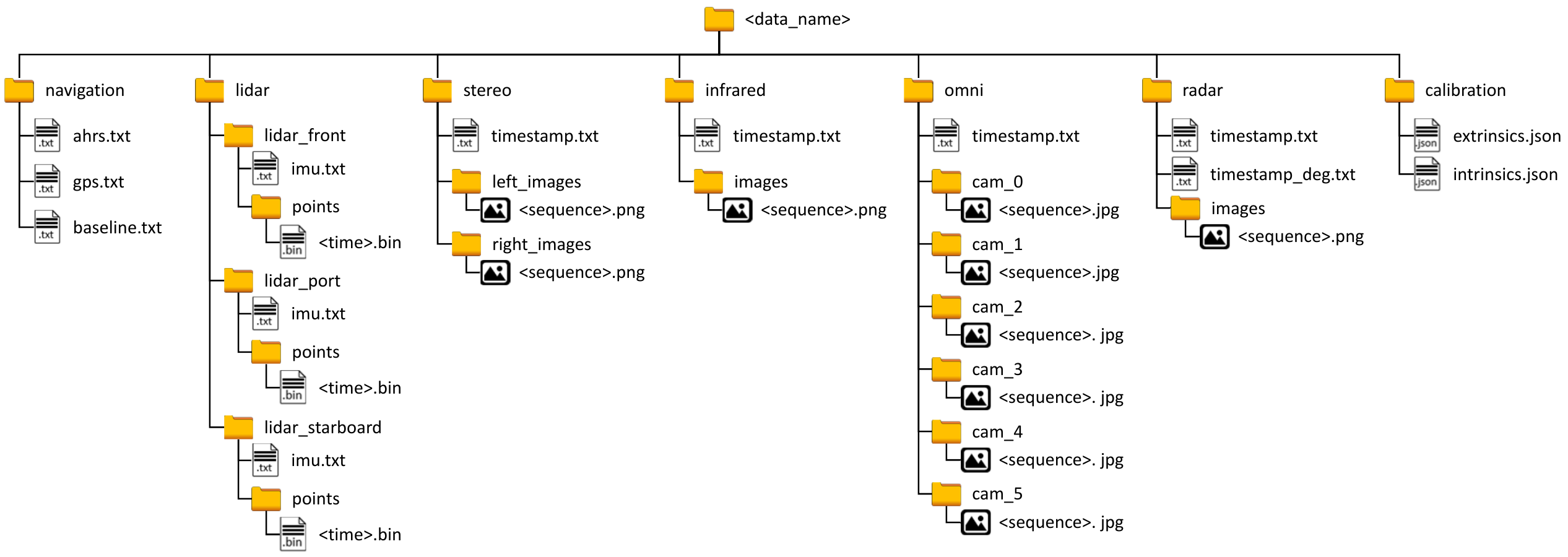}
    \caption{Dataset structure}
    \label{fig:folder_tree}
\end{figure*}

The structure of each data sequence is depicted in Fig.~\ref{fig:folder_tree}. All sensor data files were recorded along with the time they were received. The timestamps of each sensor data were saved either in the form of text files associated with the data file and its corresponding time or as data file names. The timestamps of the stereo camera, infrared camera, omnidirectional camera, and radar data were recorded as \texttt{timestamp.txt} for each sensor. These timestamp files contained the sequence of each datum (which was set as the datum name) and Unix time in the tab-delimited format. The name of each LiDAR data file was set as the timestamp in nanoseconds.

\begin{table*}[h]
\small\sf\centering
\caption{The data sequence list. Sensors are marked as L (LiDARs), R (Radar), S (Stereo Camera), I (Infrared Camera), O (Omnidirectional Camera), and N (AHRS and GPS)\label{table:data_sequence}}
\begin{tabular}{lllll}
\toprule
Data Name&Description&Path Length&Sensor Suite& Issues\\
\midrule
pohang00& Daytime& 7.43 km & L, R, S, I, O, N & GPS without RTK \\
pohang01& Nighttime & 7.46 km & L, R, S, I, N & Unstable and missing GPS data\\
pohang02& Daytime & 7.25 km & L, R, S, I, O, N & Unstable GPS data\\
pohang03& Daytime & 7.37 km & L, R, S, I, O, N & -\\
pohang04& Daytime & 7.24 km & L, R, S, I, O, N & -\\
pohang05& Nighttime & 6.86 km & L, R, S, I, N & -\\
\bottomrule

\end{tabular}
\end{table*}

\subsubsection{GPS data}

The three global navigation satellite system messages received from the GNSS-RTK receiver were GNGGA, GNHDT, and GNRMC. To record the GPS data, these three messages were combined, and eleven sequential values were saved for each timestamp in the \texttt{navigation/gps.txt} file. The values recorded in this file included Unix time, GPS time, latitude, the hemisphere of latitude (N/S), longitude, the hemisphere of longitude (E/W), heading (in degrees), GPS quality indicator, the number of satellites used, horizontal dilution of precision, and geoid height. All values were saved in the tab-delimited format.

\subsubsection{AHRS data}

The AHRS recorded orientation, angular rate, and linear acceleration data, which were saved as eleven sequential values for each timestamp in the tab-delimited format in the \texttt{navigation/ahrs.txt} file. The values included Unix time, orientation represented as quaternion values (qx, qy, qz, qw), angular rate in the x, y, and z directions, and linear acceleration in the x, y, and z directions.

\subsubsection{3D LiDAR data}

The LiDAR data from the front, port, and starboard sensors were stored in separate folders within the \texttt{lidar} directory. The front LiDAR data was stored in the \texttt{lidar\textunderscore front} folder, while the port and starboard LiDAR data was stored in the \texttt{lidar\textunderscore port} and \texttt{lidar\textunderscore starboard} folders, respectively. In each LiDAR folder, the \texttt{imu.txt} file contained inertial measurement data from the embedded IMU. The \texttt{imu.txt} file contained seven values for each timestamp, including Unix time, angular rates in the x, y, and z directions, and linear acceleration in the x, y, and z directions.

The LiDAR point cloud data was provided in binary files and included the 3D coordinates, intensity, sensor time, reflectivity, ambient, and range value of each point. The number of points in each point cloud was equal to the product of the number of channels (64 for the front LiDAR and 32 for the port and starboard LiDARs) and the axial resolution (2048 for all LiDAR data in our dataset). The data was recorded in a sequential ring-wise manner, with the ring value of each point calculated as the quotient of point sequence and the number of points in a ring. The LiDAR's internal clock recorded the sensor time of each point, which can be used to de-skew the point cloud. Each point cloud was saved in a separate file in the \texttt{points} directory with the filename format \texttt{<time>.bin}. Detailed methods for reading these binary files in both C++ and Matlab are explained on our data website.

\subsubsection{Radar data}

When the radar completed one cycle, the radar images were saved in the \texttt{radar/images} folder. As the rotation rate of the radar was low and inconsistent over time, the images were likely to be skewed whenever the platform was moved. Two timestamp files, namely \texttt{timestamp.txt} and \texttt{timestamp\textunderscore deg.txt}, were created. The \texttt{timestamp.txt} file provided only the image sequence and time pairs per cycle, while \texttt{timestamp\textunderscore deg.txt} file additionally provided the updated angle and time pairs within each image.

\subsubsection{Stereo camera data}

The left and right images from the stereo camera are stored in the \texttt{left\textunderscore images} and \texttt{right\textunderscore images} folders respectively, within the \texttt{stereo} folder. Since each stereo image pair was captured synchronously, the sequence value in \texttt{timestamp.txt} represents both the left and right image files for the same timestamp value. The images were saved in \texttt{png} format with a resolution of 2048 $\times$ 1080 pixels.

\subsubsection{Infrared camera data}

The infrared images were stored in the \texttt{infrared/images} folder. The images were recorded in 16-bit \texttt{png} format with a resolution of 640 $\times$ 512 pixels, encoding 14-bit thermal data. The temperature of each pixel can be calculated using the equation: 

\begin{equation}
    t = 0.04p - 273.15
\end{equation}

where $p$ is the pixel value and $t$ is the corresponding temperature in degrees Celsius. The data collection frequency was set to 10 Hz, but some intervals between images were longer than others due to thermal calibration.

\subsubsection{Omnidirection camera data}

The six omnidirectional camera images were stored in the \texttt{cam\textunderscore 0}, \texttt{cam\textunderscore 1}, \texttt{cam\textunderscore 2}, \texttt{cam\textunderscore 3}, \texttt{cam\textunderscore 4}, and \texttt{cam\textunderscore 5} folders within the \texttt{omni} folder. Similar to the stereo camera data, the sequence value in \texttt{timestamp.txt} indicates the corresponding images from all six cameras collected synchronously for the same time value. Each image was stored in \texttt{jpg} format with a resolution of 2464 $\times$ 2048 pixels.

\subsubsection{Calibration}

The \texttt{calibration} folder contains two files, \texttt{extrinsics.json} and \texttt{intrinsics.json}, that provide the extrinsic and intrinsic calibration parameters, respectively. The \texttt{extrinsics.json} file contains the parameters that relate the AHRS sensor to each of the other sensors. On the other hand, the \texttt{intrinsics.json} file contains the intrinsic calibration parameters for each camera, such as focal length, principal point, and distortion coefficients.

\subsubsection{Baseline trajectory}

The baseline trajectory is saved as eight sequential values for each timestamp in tab-delimited format. The values include Unix time, orientation as a quaternion (qx, qy, qz, qw), and translation (x, y, z). The trajectory data is stored in the \texttt{navigation/baseline.txt} file.

\subsection{Data sequence and known issues}

The dataset comprises six data sequences, consisting of four daytime sequences and two nighttime sequences, as shown in Table \ref{table:data_sequence}. Due to technical issues, the GPS data for sequences \texttt{pohang00} and \texttt{pohang02} was collected without RTK measurements, and sequence \texttt{pohang01} has missing GPS data during operation.

\subsection{Data player}
Inspired by \cite{ComplexUrban2019}, we provide the dataset player for ROS environment. The source code and the description can be found at https://github.com/dhchung/rosmsg\_player.

\section{Conclusion}
\label{sec:conclusion}

In this paper, we present a multimodal maritime dataset that contains continuous onboard navigation and perceptual data acquired in various maritime environments. The dataset includes GPS data obtained using a GNSS RTK receiver with two GPS antennas and AHRS data. Additionally, to address the need for various types of perceptual sensors with different characteristics and measurement ranges in the maritime environment, the dataset includes three LiDARs, a marine radar, a stereo camera, an infrared camera, and an omnidirectional camera. The dataset has been designed to facilitate full-scale autonomous navigation research and can be utilized in various domains, including sensor fusion, structure reconstruction and assessment, and other robotic applications.

\begin{acks}
This work was supported by AVIKUS corp and the city of Pohang. We greatly appreciate their support and active participation in this study.
\end{acks}

\bibliographystyle{SageH}
\bibliography{mybib.bib}
\end{document}